%% file: main.tex
\documentclass[graybox]{svmult}

\usepackage{mathptmx}       
\usepackage{helvet}         
\usepackage{courier}        
\usepackage{type1cm}        
\usepackage{makeidx}         
\usepackage{graphicx}        
\usepackage{multicol}        
\usepackage[bottom]{footmisc}
\usepackage{natbib}
\usepackage{multirow}
\usepackage{hyperref}

\usepackage{subfigure}
\usepackage{color}
\usepackage{listings}
\usepackage{caption, subcaption}
\usepackage{algorithm}
\usepackage[noend]{algpseudocode}
\usepackage[normalem]{ulem}
\usepackage{wrapfig}

\usepackage{booktabs}

\definecolor{codegreen}{rgb}{0,0.6,0}
\definecolor{codegray}{rgb}{0.5,0.5,0.5}
\definecolor{codepurple}{rgb}{0.58,0,0.82}
\definecolor{backcolour}{rgb}{0.95,0.95,0.92}

\lstdefinestyle{mystyle}{
  language=python,
  tabsize=3, 
  label=code:sample,
  frame=shadowbox,
  rulesepcolor=\color{codegray},
  xleftmargin=20pt,
  framexleftmargin=15pt,
  keywordstyle=\color{blue}\bf,
  commentstyle=\color{codegreen},
  stringstyle=\color{red},
  numbers=left,
  numberstyle=\tiny,
  numbersep=5pt, 
  breaklines=true,
  showstringspaces=false,
  basicstyle=\footnotesize,
  emph={str},
  emphstyle={\color{magenta}},
  showtabs=false
}
\makeindex             

\begin{document}

\title*{GP and LLMs for Program Synthesis: No Clear Winners} 
\titlerunning{GP and LLMs for Program Synthesis: No Clear Winners}
\author{Jose Guadalupe Hernandez, Anil Kumar Saini, Gabriel Ketron,  Jason H. Moore}
\authorrunning{Hernandez, Saini, Ketron, Moore}
\institute{
Jose Guadalupe Hernandez \at Cedars-Sinai Medical Center, Los Angeles, CA, USA \email{jose.hernandez8@cshs.org} \and 
Anil Kumar Saini \at Cedars-Sinai Medical Center, Los Angeles, CA, USA \email{anil.saini@cshs.org} \and
Gabriel Ketron \at Cedars-Sinai Medical Center, Los Angeles, CA, USA \email{gabriel.ketron@cshs.org} \and
Jason H. Moore \at Cedars-Sinai Medical Center, Los Angeles, CA, USA \email{jason.moore@csmc.edu}
}
%
%
\maketitle

\abstract*{}

\abstract{ Genetic programming (GP) and large language models (LLMs) differ in how program specifications are provided: GP uses input-output examples, and LLMs use text descriptions.
In this work, we compared the ability of PushGP and GPT-4o to synthesize computer programs for tasks from the PSB2 benchmark suite.
We used three prompt variants with GPT-4o: input-output examples (data-only), textual description of the task (text-only), and a combination of both textual descriptions and input-output examples (data-text).
Additionally, we varied the number of input-output examples available for building programs.
For each synthesizer and task combination, we compared success rates across all program synthesizers, as well as the similarity between successful GPT-4o synthesized programs.
We found that the combination of PushGP and GPT-4o with data-text prompting led to the greatest number of tasks solved (23 of the 25 tasks), even though several tasks were solved exclusively by only one of the two synthesizers.
We also observed that PushGP and GPT-4o with data-only prompting solved fewer tasks with the decrease in the training set size, while the remaining synthesizers saw no decrease.
We also detected significant differences in similarity between the successful programs synthesized for GPT-4o with text-only and data-only prompting.
With there being no dominant program synthesizer, this work highlights the importance of different optimization techniques used by PushGP and LLMs to synthesize programs.
}

\input{Text/introduciton}
\input{Text/motivation}
\input{Text/background}
\input{Text/related-works}
\input{Text/methods}
\input{Text/results}
\input{Text/conclusion}

\section*{Acknowledgments}

We thank the Department of Computational Biomedicine at Cedars-Sinai Medical Center for their feedback and for providing High-Performance Computing resources.
Cedars-Sinai Medical Center provided computational resources through their High Performance Computing clusters.
The work was supported by NIH grants R01 LM010098, U01 AG066833, and R01 LM014572.

\bibliographystyle{apalike}
\bibliography{references,software}

\end{document}

%% file: Text/introduciton.tex
 \section{Introduction}


Program synthesis approaches attempt to automatically construct computer programs that satisfy a set of user-provided specifications, which define the expected behavior of a correct solution \citep{gulwani2017program}.
Recent advancements in large language models (LLMs) have increased interest in program synthesis by enabling the specification of a program’s intended behavior through textual descriptions, for which the LLM subsequently outputs executable code.
LLMs are increasingly being adapted for a broad spectrum of programming tasks, such as debugging \citep{chen2023teaching}, code summarization \citep{ahmed2024summarization}, software testing \citep{wang2024testing}, and comprehensive end-to-end software engineering workflows \citep{yang2024swe}.
Although LLMs are emerging as leading tools in software development and program synthesis, the quality of the generated programs remains contingent on the specificity and clarity of the provided descriptions \citep{chen2021evaluatinglargelanguagemodels}.


\textit{User intent} refers to the user-defined specifications or desired behaviors that a program is expected to exhibit. 
These intentions can be expressed through various modalities, such as input-output examples, natural language descriptions, or partially completed code fragments \citep{gulwani2017program}.
Program synthesis techniques that interpret user intent through input-output examples are commonly categorized as Inductive Program Synthesis or Programming by Example \citep{polozov2015flashmeta, gulwani2017program}.
Specifying input-output examples is one of the most accessible approaches for both novice and experienced programmers, as it allows users to define the intended program behavior without requiring an explicit understanding of the underlying implementation that maps inputs to their corresponding outputs.
This approach is particularly valuable for synthesizing solutions to complex or novel programming tasks, especially in cases where explicitly specifying an implementation---such as devising an efficient algorithm for an NP-hard problem---is either unintuitive or computationally intractable.


Regardless of how user intent is specified, a program synthesizer must generate a program that satisfies the given requirements.
Finding the optimal program, however, is challenging due to the vast space of potential candidates to consider.
Various approaches constrain this search space by restricting the set of possible programs, such as synthesizers that operate within a Domain-Specific Language~\citep{perelman2014test}.
In contrast, synthesizers utilizing Turing-complete languages can theoretically produce any computable function, drastically increasing the size and complexity of the search space.
LLMs employ a probabilistic approach, guided by extensive training on textual data that includes examples of existing programs and software \citep{openai2024gpt4technicalreport,chen2021evaluatinglargelanguagemodels,touvron2023llama}.
This training approach enables LLMs to iteratively generate text for synthesizing functional programs by predicting the most likely continuation at each step.
Genetic programming (GP), however, typically constructs programs from scratch, evaluating them for correct behavior throughout the evolutionary search process.


In this work, we compare the effectiveness of PushGP and GPT-4o in generating solutions for program synthesis tasks from the PSB2 benchmark suite \citep{helmuth2021psb2}, extending and building upon findings presented by \cite{hernandez2024comparing}.
Specifically, we evaluate three novel experimental configurations:
(1) varying the number of input-output cases supplied to each program synthesizer;
(2) assessing the similarity among optimal programs;
and (3) introducing a hybrid prompt for GPT-4o, which integrates elements from the data-only and text-only prompts described by \cite{hernandez2024comparing}.
Ultimately, no single program synthesizer consistently outperforms the others across all experimental settings.
However, under specific conditions, a partial ordering of synthesizer preferences emerges, which can inform the selection of tools for successful program generation.

%% file: Text/motivation.tex
\section{Motivation: Fair Comparisons Between Genetic Programming and Large Language Models for Program Synthesis}


In \cite{sobania2024comparison,sobania2022choose}, the authors compared the effectiveness of GitHub Copilot and genetic programming (GP) for program synthesis.
Copilot received prompts that included both a function signature and a textual task description derived from the tasks in PSB1 \citep{helmuth2015psb1} and PSB2 \citep{helmuth2021psb2} suites, two widely used benchmarks in GP-based program synthesis.
The results for GP were derived from prior studies that used a variety of GP systems \citep{helmuth2022problem,forstenlechner2018extending,boldi2024informed,sobania2021recent}.
Both studies found that GP and Copilot successfully solved a comparable number of tasks.
However, Copilot demonstrated a higher success rate and produced more user-friendly programs across multiple tasks.
We believe that Copilot’s success over GP stems from fundamental differences in their intended use cases and the distinct user intent format each synthesizer employs to address a given problem.


Users with an understanding of a program’s intended functionality and design can enhance the quality of results generated by a large language model (LLM) through prompt engineering, which involves refining prompts with detailed descriptions and relevant test cases \citep{shin2023prompt,murr2023testing}.
In practice, however, users may not always have access to a detailed specification on how a program should be designed or implemented.
For example, consider a scenario in which a program is being developed using a test-driven development approach; unit tests specifying the program’s expected behavior are written before implementation.
Depending on the program’s complexity, both expert and novice programmers may encounter challenges, but the unit tests can help all programmers verify if a program is behaving correctly.
We believe GP thrives in these specific scenarios, where input-output examples are the primary source guiding the generation of computer programs.

Copilot and GP both address the same program synthesis tasks in \cite{sobania2024comparison,sobania2022choose}, but differ the format that user intent is provided to each synthesizer.
Copilot receives prompts with details on how programs should operate, derived from the textual descriptions in PSB1 and PSB2, while GP relies soley on input-output examples to infer program behavior.
This difference in user intent results in both program synthesizers working with different types of information for the same task, even though both synthesizers are attempting to solve the same task.
For example, in the Middle Character task from PSB2, the description states: ``Given a string, return the middle character as a string if it is odd length; return the two middle characters as a string if it is even length.'' 
We believe this explicit instruction allows Copilot to leverage its knowledge of the programs in its training corpus more effectively.
Naturally, if a detailed description of how a program should function is available, an LLM-based synthesizer should be preferred.
However, how would LLM-based program synthesizers perform compared to GP if both were limited to input-output examples as the basis of user intent?
\cite{hernandez2024comparing} investigated this question and found that GP successfully solves a greater number of problems from the PSB2 benchmark compared to GPT-4o, with GPT-4o primarily excelling on problems frequently encountered online.


GitHub Copilot is a descendant of Codex \citep{chen2021evaluatinglargelanguagemodels}, an OpenAI GPT-3 model with approximately 12 billion parameters, trained on 159 GB of code sourced from 54 million public GitHub repositories.
We can reasonably assume that Copilot's training processes spanned several months, based on the estimated \$4.3 million training cost and computational requirement exceeding 100 million petaFLOPS for the GPT-3 foundational model \citep{aiindex}.
While the construction of Copilot required significant time and resources, its model inference is nearly instantaneous, providing users with code suggestions and assistance within seconds.
In contrast, GP systems require more time to produce a program, as they must perform an evolutionary search and evaluate candidate programs.
While the immediate output of an LLM may appear advantageous, it lacks the iterative validation inherent in GP, where candidate programs are evaluated and refined through the search process.
GP systems are typically not granted extensive computational time or resources like Copilot.
As such, Copilot has a significant advantage in terms of total computational capacity when considering both training time and model inference.

We recommend the following three factors to facilitate fairer comparisons between GP and LLMs:
\begin{enumerate}
    \item \textbf{Same User Intent:} The user intent for either GP and LLMs capture the same information found within the problem statement.
    \item \textbf{Equivalent Attempts Per Run:} Limit LLMs to a single, independent attempt per task such that both GP and LLMs returns only one solution.
    \item \textbf{Equal Resource Allocation:} Increasing the number of evaluated programs in a GP run to match Copilot's total compute used for training and inference time.
\end{enumerate}
In this work, we compare the program synthesis capabilities of GPT-4o \citep{openai2024gpt4technicalreport} and PushGP, adhering to the first and second principles listed above, leaving the other for future work.

%% file: Text/background.tex
\section{Background}

\subsection{Program synthesis}


Automatically synthesizing computer programs is a key goal in the pursuit of Artificial General Intelligence \citep{oneill2020automatic} and is often regarded as the ``holy grail'' of computer science \citep{gulwani2017program}.
Indeed, the ability to automatically generate all programs would benefit anyone interested in programming \citep{david2017program}, as computers are becoming increasingly accessible.
Contemporary program synthesizers serve multiple roles: 
they enable non-programmers to generate programs from textual descriptions ~\citep{sharlin2024contextlearningreasoningsymbolic}, 
assist programmers in implementing auxiliary functions \citep{gulwani2010dimensions}, 
and discover programs and designs that match or surpass the quality of those made by human programmers \citep{kannappan2015analyzing}.


A program synthesizer can be categorized by how it accepts and processes user intent, the space of all possible programs, and the optimization strategy used to produce programs \citep{gulwani2010dimensions,gulwani2017program}.
These three characteristics create distinct challenges that effect program synthesis success.
For example, solving a program synthesis task requires the program space to encompass the correct programs and the optimization strategy to be sufficiently robust to find those programs.
Fortunately, users pick the program synthesizer they wish to use, giving them control over the choice of the program space and the optimization strategy.
User intent is meant to capture the specifications that define the behavior of a correct program, but articulating this intent often represents the most challenging aspect of program synthesis.


Deductive synthesis is one of the earliest approaches for program synthesis, leveraging automated theorem provers and typically relying on precise user-provided specifications to generate programs \citep{green1969application,manna1971toward,waldinger1969prow}.
However, providing a precise set of specifications to guide the synthesis of correct programs has proven challenging, as it requires expertise in the specific problem domain. 
Even with sufficient domain knowledge, mitigating ambiguity within user intent remains challenging  \citep{gulwani2017program,oneill2020automatic}.
Inductive synthesis provides an alternative approach, where user intent can be expressed less comprehensively, often through examples.
Programmers may find input-output examples more intuitive, as they mimic unit tests commonly used to verify program correctness.

\subsection{Large Language Models (LLMs)}

Large Language Models (LLMs) are neural network architectures that produce output sequences by predicting the elements that should follow a given input sequence.
The most advanced LLMs use a Generative Pre-trained Transformer (GPT) architecture.
In the  paper, ``Attention Is All You Need'', \cite{vaswani2023attentionneed} demonstrated that the Transformer model could wield massive improvements to training speed and next-word prediction quality by relying on the attention mechanism. 
Attention allows language models to weight the relative importance of each word in a sentence against all the other words of the sentence simultaneously, and later using said weighting to predict the most likely next word.

%% file: Text/related-works.tex
\section{Related Work}

\subsection{GP for program synthesis}


The use of genetic programming (GP) for program synthesis has experienced a significant rise in the number of published works \citep{sobania2023comprehensive}.
Three GP techniques are typically used:
Stack-based GP \citep{perkis1994stacks}, where programs are represented as stacks containing instructions that manipulate data; 
Grammar-guided GP \citep{whigham1995GBGP}, where programs are generated according to user-defined context-free grammars, ensuring that only syntactically valid programs are produced; 
Linear GP \citep{brameier2007lgp}, where programs consist of a linear sequence of instructions that operate on variables and interact with memory registers.
In this work, we use PushGP \citep{spector2005push3}, a GP system that evolves programs represented as linear sequences of datatype-specific instructions and constants.
The Program Synthesis Benchmark Suites provide two benchmark suites, PSB1 \citep{helmuth2015psb1} and PSB2 \citep{helmuth2021psb2}, that serve as the standard benchmarks for GP-based program synthesis.


In \cite{sobania2023comprehensive}, the authors provide a comprehensive survey of the performance of various GP systems on all 29 tasks of the PSB1 suite.
The results are derived from 54 papers that report the success of specific GP systems on these tasks.
Interestingly, problem-solving success was observed for all but three problems in the PSB1 benchmark suite, with notable differences in performance across different GP systems.
Lexicase \citep{helmuth2014solving} has improved the program synthesis capabilities of some GP systems \citep{helmuth2020benchmarking}. 
Instead of aggregating the errors on multiple test cases, lexicase uses individual error values to select parents. 
In Helmuth \& Spector~\citep{helmuth2022problem}, the authors reported that standard lexicase selection successfully solved 13 of the 25 tasks from the PSB2 suite, whereas down-sampled lexicase \citep{hernandez2019random}, a lexicase variant, demonstrated superior performance, solving 17 tasks.

\subsection{LLMs for Program Synthesis}


The capabilities of large language models (LLMs) have expanded the scope of program synthesis by allowing user intent to be specified through natural language and introducing a new program search mechanism that predicts the next word in a program.
LLMs used for program synthesis are classified as inductive synthesis approaches because they can receive program descriptions or input-output examples. 
Interestingly, studies of LLM-based program synthesis typically use natural language descriptions for user intent \citep{Ramirez-Rueda_Benitez-Guerrero_Mezura-Godoy_Bárcenas_2024}, whereas input-output examples are less frequent~\citep{jigsaw, murr2023testing}.

OpenAI's GPT-4, and its cost-effective variant GPT-4o, extends earlier decoder-transformer LLM architectures by incorporating image processing capabilities and expanding its parameter space to enhance performance and versatility \citep{openai2024gpt4technicalreport}.
GPT-4 achieved success on 67\% of program synthesis tasks from the HumanEval benchmark, a dataset comprising 164 programming problems. 
Each task includes a prompt containing a function signature, a docstring (a natural language description embedded within Python functions), and a set of unit tests.
Additionally, GPT-4 also solved 31 of the 41 tasks classified as `easy' on LeetCode \citep{openai2024gpt4technicalreport}, an online platform hosting a wide range of coding problems designed for programming competitions and technical interviews.

There are also studies that aim to analyze the impact of different prompt styles on program synthesis performance.
For example, \cite{murr2023testing} analyzed the success rates of different types of user intent in program synthesis, where each type represents the same prompt rephrased or structured distinctly.
Specifically, they evaluated prompts structured in four ways: textual descriptions alone, textual descriptions combined with test cases, test cases alone, and generic function prompts with test cases (i.e., no hints were provided in the function signature).
Interestingly, they found that generic function prompts with test cases solved the fewest number of problems, with GPT-4 being the best model tested for writing successful code with these kinds of prompts.
Note that these generic function prompts closely resemble the ones used in this work.

%% file: Text/methods.tex
\section{Methods}

We compare the program synthesis capabilities of PushGP and GPT-4o.
For GPT-4o, three prompt variants are used to specify program synthesis tasks: prompts consisting solely of input-output examples (data-only), prompts comprising only textual task descriptions (text-only), and prompts that combine both input-output examples and textual descriptions (data-text).
PushGP is only provided a set of input-output examples.
This configuration effectively creates four program synthesizers, which are evaluated on tasks from the Program Synthesis Benchmark 2 (PSB2) \citep{helmuth2021psb2} suite.
Each of task is defined by a textual description and a corresponding set of input-output training cases.
Additionally, we vary the number of training cases (50 and 200) provided to each program synthesizer to assess the effect of training set size on performance.


For each of the 25 tasks from the PSB2 suite, we evaluate eight different experimental conditions: four program synthesizers in combination with two training set sizes.
For each combination of task, synthesizer, and training set size, we performed 100 replicates:
specifically, 100 PushGP runs and 100 queries to GPT-4o for each prompt variant (three total).
A replicate is deemed successful if it produces a program that satisfies all examples in both the training and test sets.
We evaluate the similarity among successful programs generated by GPT-4o for a specific task to assess how consistently each GPT-4o synthesizer produces similar solutions.
Note that GPT-4o was prompted to generate Python functions and PushGP produced programs in the Push programming language.

\subsection{Program Synthesis Benchmark 2 (PSB2)}
\label{sec:benchmark}

The PSB2 \citep{helmuth2021psb2} suite comprises 25 tasks of moderate difficulty, drawn from coding competitions and programming courses.
Each task is defined by a textual description and a corresponding set of input-output examples.
The tasks are derived from four sources:
\begin{itemize}
    \item Code Wars: User-created coding challenges hosted online.
    \item Advent of Code: Coding problems hosted online for training, interview preparation, or coursework.
    \item Project Euler: Online archive of hundreds of problems.
    \item Homework Problems: Coding homework problems given in undergraduate programming courses (not online).
\end{itemize}

The set of input-output examples for each task comprises one million randomly generated examples that conform to the task’s constraints and a small set of examples designed to address edge cases.
The ``training'' set first includes edge cases for each task, with the rest sampled from the one million examples.
The test set consists of 2,000 randomly sampled examples from the one million examples.
Here, we use 200 and 50 training examples to guide program synthesizers that rely on input-output examples. 
Note that if a given task includes 50 or more edge cases, the training set of size 50 may consist exclusively of such cases.

For PushGP, new training and test sets are sampled each run. 
It is important to note that this approach is recommended in the PSB2 paper \citep{helmuth2021psb2} and has been implemented online\footnote{https://github.com/thelmuth/psb2-clojure/blob/main/src/psb2/core.clj}. 
For all GPT-4o runs, the same 100 training and test splits are used across each prompt variant, with variation introduced by using a temperature of $0.7$ and altering the random state each time GPT-4o is queried.
Additionally, both sets of training and test cases were resampled for both training set size conditions.

\subsection{PushGP Configuration}
\label{sec:pushgp}

PushGP is written in the Push programming language, and we use its Clojure-based implementation, known as Clojush\footnote{https://github.com/lspector/Clojush}.
For each PSB2 task, we execute 100 independent Clojush runs, with training and test sets resampled for each run.
We used the genetic parameters that have shown the maximum performance on PSB1 and/or PSB2 problems: downsampled-lexicase~\citep{hernandez2019random} as the parent selection method, and Uniform Mutation by Addition and Deletion (UMAD) \citep{helmuth2018program} as the mutation operator.
We use a population size of 1000, and the genetic operator must construct the same number of offspring every generation.
A mutation rate of 0.09 is used, and no crossover is applied in this setting.
As recommended in \cite{helmuth2022problem}, we use the down-sample size of 0.25 and 1200 generations.
Our supplemental material [tbd] contains full details on which problems used which genetic operator settings.

\begin{figure}[ht!]
    \captionof{figure}{Data-only prompt for the `Spin-Words' task.}
    \begin{lstlisting}
```python
def my_func(input1:str):
"""Alter this python function "my_func" to accept inputs 
containing a string of length [0, 20]. The function should 
output a string that replicates the underlying mechanism of 
the following examples. Only use base python functions and do 
not import any packages. Do not include print statements,
unit tests, in-line comments or multi-line comments.
Examples: my_func('helpful') == 'lufpleh'
...
"""
```
    \end{lstlisting}
    \label{fig:D} 
    \captionof{figure}{Text-only prompt for the `Spin-Words' task.}
    \begin{lstlisting}
```python
'Given a string of one or more words (separated by spaces), 
reverse all of the words that are five or more letters long 
and return the resulting string.'
def my_func(input1:str):
"""Alter this python function "my_func" to accept inputs 
containing a string of length [0, 20]. The function should 
output a string. Only use base python functions and do not 
import any packages. Do not include print statements, unit 
tests, in-line comments or multi-line comments."""
```
    \end{lstlisting}
    \label{fig:T} 
    \captionof{figure}{Data and text prompt for the `Spin-Words' task.}
    \begin{lstlisting}
```python
'Given a string of one or more words (separated by spaces), 
reverse all of the words that are five or more letters long 
and return the resulting string.'
def my_func(input1:str):
"""Alter this python function "my_func" to accept inputs 
containing a string of length [0, 20]. The function should 
output a string that replicates the underlying mechanism of 
the following examples. Only use base python functions
and do not import any packages. Do not include print 
statements,unit tests, in-line comments or multi-line 
comments. 
Examples: my_func('helpful') == 'lufpleh'
...
"""
```
    \end{lstlisting}
    \label{fig:DT}
\end{figure}
\subsection{OpenAI's GPT-4o Configuration}


We evaluated the program synthesis capabilities of GPT-4o, the publicly available model accessible via Azure’s cost-efficient proprietary OpenAI API \citep{gpt-4o}.
Specifically, we used the Azure OpenAI Service GPT-4o version dated 2024-08-06, configured with a limit of 990,000 tokens per minute, a temperature of $0.7$, and random state altered each time GPT-4o is queried.
Details regarding the use and configuration of GPT-4o are provided in the supplemental material.

The structure and quality of the prompt provided to a large language model (LLM) can significantly influence the quality of the synthesized program ~\citep{chen2021evaluatinglargelanguagemodels}.
Therefore, we reviewed the literature to identify prompt structures that emphasize the effective processing of input-output examples for program synthesis.
Here, we adopt the prompt identified by \cite{hernandez2024comparing}, where the authors evaluated several prompt variants and determined that the one used in this work yielded the best performance for GPT-4o.

We evaluated GPT-4o using three variants of the optimal prompt: data-only prompts, text-only prompts, and data-text prompting.
See Figures 1, 2, and 3 for examples of these prompts for the Spin Words task.
Data-only prompts ensure that both PushGP and GPT-4o receive equivalent information for program synthesis, while text-only prompts align with the conventional approach prompts are typically structured for LLMs.
The combination of both assesses the effect of both styles integrated within a single prompt.
We conducted 100 independent API calls for each combination of PSB2 task, training set size, and prompt type used with GPT-4o.
For each query, the synthesized Python function was evaluated against all training examples for a given task.
Solutions that correctly solved both the training and test sets constituted a successful run. 

\subsection{Data Tracking \& Statistical Analysis}
\label{methods:stats}

We use \textit{success rate} to measure the program synthesis capabilities of the synthesizers used in this study.
This metric is defined as the number of runs (for PushGP) or queries (for GPT-4o) out of 100 that successfully generated a program solving all examples in both the training and test sets.
We provide the success rates for PushGP and GPT-4o on the benchmark problems in Table \ref{tab:succ_rate_200} for 200 training cases, and success rates for 50 training cases in Table \ref{tab:succ_rate_50}.

We use a Pairwise Chi-squared test with $\alpha = 0.05$ and Holm correction to compare the success rates across PSB2 tasks among the four methods studied here (See Tables \ref{tab:succ_rate_200} and \ref{tab:succ_rate_50}).
Specifically, for all four values (success rates) for a given problem, we measure the statistical significance of all pairwise differences. 
The best success rates per task are underlined in Tables \ref{tab:succ_rate_200} and \ref{tab:succ_rate_50}.

To compute similarity, we use an open-source\footnote{https://github.com/blingenf/copydetect} implementation of the Winnowing algorithm \citep{schleimer2003winnowing}. 
This algorithm begins by preprocessing the source code: Python programs are tokenized and filtered.
The similarity computation proceeds by generating $k$-grams from the preprocessed code, hashing each $k$-gram, and then applying a sliding window of size $w$ over the resulting hashes. 
Within each window, the minimum hash value is selected as a fingerprint.
The similarity from one program to anther is calculated as the ratio of overlapping fingerprints to the total number of fingerprints.
The parameters $k$ and $w$ influence the sensitivity of the algorithm \citep{shrestha2023winnowing,astica2023analysis}; higher values tend to decrease sensitivity to small changes. 
We evaluate the algorithm using three settings for $k$ and $w$---2, 5, and 10---where $k = w$ in each case.

For each GPT-4o synthesizer and task combination with at least two successful replicates, we first compute the pairwise similarity between each successful program and all others, averaging these values to obtain a similarity score for each individual program. 
These individual scores are then averaged across all successful programs for the task to produce a single representative similarity value.
We use a Kruskal-Wallis test with an $\alpha=0.05$ to check for significant differences among all three GPT-4o prompt variants.
For comparisons where significant differences were detected, we used a Wilcoxon Rank Sum test with with a Bonferroni correction for multiple comparisons and an $\alpha=0.05$ to identify significant differences among pairwise comparisons.

\subsection{Software availability}

Our supplemental material \citep{supplemental_material} is hosted on \href{https://github.com/}{GitHub} and contains all the software, data analysis, and documentation for this work.
Our experiments are implemented in Python and Clojure.
We also used a combination of Python 3.10 and R version 4 for data collection, visualization, and analysis.
All data generated for this work can be found at \href{https://osf.io/s7vhj/}{https://osf.io/s7vhj/}.

%% file: Text/results.tex
\section{Results \& Discussion}
\subsection{Results for success rates with 200 training cases}

\subsubsection{PushGP and GTP-4o with data-text prompts solve similar tasks but at different success rates}
\label{res:push_gpt-dt}


As shown in Table \ref{tab:succ_rate_200}, both PushGP and GPT-4o with data-text prompts successfully solved more tasks than the other two program synthesizers on the PSB2 benchmark suite, solving 17 and 18 tasks, respectively (the first and third columns in Table \ref{tab:succ_rate_200}).
Of the 18 tasks successfully solved by GPT-4o with data-text prompts, PushGP was able to produce at least one successful program on 12 tasks: Basement, Camel Case, Coin Sums, Find Pair, Fizz Buzz, Fuel Cost, GCD, Middle Character, Paired Digits, Solve Boolean, Substitution Cipher, and Twitter.
This highlights a substantial overlap in the tasks successfully addressed by both program synthesizers.
Among this overlap, however, GPT-4o managed to significantly outperform PushGP in 10 of the tasks (Chi-squared test: $p<10^{-3}$), whereas PushGP significantly outperformed GPT-4o on the Twitter task (Chi-squared test: $p<10^{-3}$).
No significant differences were detected on the Solve Boolean task (Chi-squared test: $p>0.05$).


PushGP failed to solve the six remaining tasks that GPT-4o successfully completed using data-text prompts: Bowling, Luhn, Mastermind, Shopping List, Spin Words, and Square Digits.
Among these six tasks, PushGP was significantly outperformed on all of these tasks (Chi-squared test: $p<10^{-1}$).
Conversely, no GPT-4o configuration managed to solve the remaining five tasks that PushGP managed to solve: Bouncing Balls, Cut Vector, Dice Game, Indices of Substring, and Snow Day.
Among these five tasks, all GPT-4o configurations were significantly outperformed on the Snow Day task (Chi-squared test: $p<10^{-2}$) and no significant differences were detected on all remaining tasks (Chi-squared test: $p>0.05$).


These results suggest that when a task can be solved by both PushGP and GPT-4o using data-text prompts, GPT-4o should be preferable given its significantly better success rate across multiple problems.
However, the use of data-text prompting assumes the necessary information to construct the prompt is available.
If GPT-4o fails to solve a task despite the use of data-text prompting, PushGP presents a viable alternative for program synthesis, as it successfully generated solutions for five tasks that no other synthesizer was able to solve.
Ultimately, the combined use of both program synthesizers proves effective, collectively solving 23 out of the 25 available tasks.

\begin{table}[ht!]
\centering
  \caption{Success Rates for PSB2 tasks for PushGP and GPT-4o when provided 200 training cases. 
  The underlined numbers indicate the highest success rates in a given row. 
  (D) represents data-only prompts, (T) represents text-only prompts, and (D-T) represents the combination of both prompts.
  The last row shows the number of tasks solved by a particular method that were not solved by other methods.
  We discuss specific p-values of pairwise comparisons in the text.}
  \label{tab:succ_rate_200}
  \begin{tabular}{l|c|c|c|c}
    \toprule
     Task & PushGP &  GPT-4o (D)  & GPT-4o (D-T) & GPT-4o (T)\\
       \midrule
        Basement & 1 & 0 & \underline{100} & \underline{100} \\
        Bouncing Balls & \underline{3} & 0 & 0 & 0 \\
        Bowling & 0 & 13 & \underline{19} & 12 \\
        Camel Case & 1 & 95 & \underline{100} & \underline{100} \\
        Coin Sums & 17 & 0 & \underline{76} & 0 \\
        Cut Vector & \underline{1} & 0 & 0 & 0 \\
        Dice Game & \underline{3} & 0 & 0 & 0 \\
        Find Pair & 19 & 49 & 99 & \underline{100} \\
        Fizz Buzz & 82 & \underline{100} & \underline{100} & \underline{100} \\
        Fuel Cost & 56 & 2 & \underline{100} & \underline{100} \\
        GCD & 14 & 95 & \underline{100} & \underline{100} \\
        Indices of Substring & \underline{1} & 0 & 0 & 0 \\
        Leaders & 0 & 0 & 0 & 0 \\
        Luhn & 0 & 0 & \underline{7} & 0 \\
        Mastermind & 0 & 38 & \underline{80} & 0 \\
        Middle Character & 75 & 9 & 99 & \underline{100} \\
        Paired Digits & 18 & 4 & \underline{91} & 69 \\
        Shopping List & 0 & 1 & \underline{53} & 0 \\
        Snow Day & \underline{12} & 0 & 0 & 0 \\
        Solve Boolean & 6 & 3 & \underline{17} & 14 \\
        Spin Words & 0 & 52 & \underline{100} & \underline{100} \\
        Square Digits & 0 & 70 & \underline{100} & 99 \\
        Substitution Cipher & 77 & 29 & 97 & \underline{100} \\
        Twitter & \underline{56} & 11 & 11 & 0 \\
        Vector Distance & 0 & 0 & 0 & 0 \\
        \midrule
        Num. of Tasks Solved & 17 & 15 & 18 & 13\\
        \midrule
        Num. of Unique Tasks Solved & 5 & 0 & 1 & 0 \\
  \bottomrule
\end{tabular} 
\end{table} 

\subsubsection{PushGP solves more tasks when only input-output cases are available}

Although PushGP and GPT-4o interpret user intent through distinct representational paradigms, this intent can be encoded to preserve equivalent information. 
Here, GPT-4o with data-only prompts ensures that both synthesizers are tasked with generating programs that solve the same input-output mapping problem.
In Table \ref{tab:succ_rate_200}, we observed that PushGP evolved successful programs for 17 of the 25 PSB2 tasks, whereas GPT-4o with data-only prompts generated successful programs for only 15 tasks (the first and second columns in Table \ref{tab:succ_rate_200}).
Interestingly, both of these program synthesizers successfully solved 10 overlapping tasks: Camel Case, Find Pair, Fizz Buzz, Fuel Cost, GCD, Middle Character, Paired Digits, Solve Boolean, Substitution Cipher, and Twitter.
Among these overlapping tasks, no significant differences were detected on the Solve Boolean task (Chi-squared test: $p>0.05$), while GPT-4o significantly outperformed PushGP on four tasks (Chi-squared test for Camel Case, Find Pairs, Fizz Buzz, and GCD: $p<10^{-3}$).
For the remaining five tasks, PushGP significantly outperformed GPT-4o (Chi-squared test for Fuel Cost, Middle Character, Paired Digits, Substitution Cipher, and Twitter: $p<10^{-2}$).

GPT-4o with data-only prompts managed to successfully solve five tasks that PushGP could not: Bowling, Mastermind, Shopping List, Spin Words, and Square Digits.
Interestingly, no significant differences in success rate were detected for the Shopping List task (Chi-squared test: $p>0.05$), while GPT-4o significantly outperformed PushGP on all remaining tasks (Chi-squared test: $p<10^{-2}$).
Conversely, PushGP successfully solved seven task that GPT-4o with data-only prompts could not: Basement, Bouncing Balls, Coin Sums, Cut Vector, Dice Game, Indices of Substring, and Snow Day.
Of these seven tasks, PushGP significantly outperformed GPT-4o on the Coin Sum and Snow Day tasks (Chi-squared test: $p<10^{-2}$), while no significant differences were detected for the remaining tasks (Chi-squared test: $p>0.05$).

Additionally, we found that GPT-4o with data-only prompts produced successful programs for fewer tasks compared to GitHub Copilot: for their Copilot system, \cite{sobania2024comparison} reported success on 20 of the 25 PSB2 tasks, and \cite{sobania2022choose} reported success on 15 tasks.
We suspect that the discrepancies between our GPT-4o results and Copilot are primarily due to differences in prompt modalities and the underlying LLM architectures: Copilot receives structured prompts containing task descriptions and is explicitly optimized for programming-related tasks.
Our GPT-4o with data-only prompt findings also diverge from those reported in \cite{hernandez2024comparing}, likely due to the data preprocessing strategies employed in this work, which are tailored to specific data types to guide the LLM toward processing more semantically appropriate input.
The improvements for PushGP can be attributed to the use of down-sampled lexicase, as it has been previously shown to excel at program synthesis compared to other parent selection algorithms, especially on the PSB2 suite.


Both program synthesizers appear equally matched for program synthesis when only input-output examples are available. 
GPT-4o with data-only prompting may be more beneficial for users that require an instant solution given the almost instantaneous inference time, especially given the success rate improvements on overlapping and non-overlapping tasks.
However, PushGP also presents a worthy approach as it manages to solve more overall and unique tasks as compared to GPT-4o with data-only prompting.
Ultimately, the preferred approach depends on the needs of the user, where problems that may not be solved by GPT-4o warrant further investigation by PushGP.
This suggestion is evident by the fact that PushGP and GPT-4o with data-only prompting successfully solved 22 tasks---just one fewer than the number of joint successes reported in Section \ref{res:push_gpt-dt} for PushGP and GPT-4o with data-text prompting.

\subsubsection{Among the three prompt variants, GPT-4o with data-text prompts solves the most tasks}

The effectiveness of prompt-based program synthesis with LLMs is highly dependent on prompt quality; however, prompt quality itself is constrained by the amount and relevance of information that can be incorporated into the prompt.
The PSB2 suite provides both input-output examples and natural language task descriptions, enabling us to isolate the individual contributions of each component as well as their combined effect on synthesis performance of GPT-4o.
In Table \ref{tab:succ_rate_200}, data-text prompting resulted in the highest number of successfully solved tasks, followed by data-only and text-only prompting (second, third, and fourth column in Table \ref{tab:succ_rate_200}). 
Specifically, data-text prompting solved 18 tasks, while data-only and text-only prompting solved 15 and 13 tasks, respectively.
The tasks solved by data-only and text-only prompting completely overlap with the tasks solved only by data-text prompting.

The data-text and data-only prompting approaches overlap on 15 tasks: Bowling, Camel Case, Find Pair, Fizz Buzz, Fuel Cost, GCD, Mastermind, Middle Character, Paired Digits, Shopping List, Solve Boolean, Spin Words, Square Digits, Substitution Cipher, and Twitter.
Among these overlapping tasks, no significant differences between prompt types were detected for the Bowling, Camel Case, Fizz Buzz, GCD, and Twitter tasks (Chi-squared test: $p>0.05$).
For the remaining ten tasks, data-text prompting significantly outperformed data-only prompting (Chi-squared test: $p<10^{-1}$).

The data-text and text-only prompting approaches overlap on 13 tasks: Basement, Bowling, Camel Case, Find Pair, Fizz Buzz, Fuel Cost, GCD, Middle Character, Paired Digits, Solve Boolean, Spin Words, Square Digits, and Substitution Cipher.
Among these overlapping tasks, data-text prompting outperformed text-only prompting on the Paired Digits task (Chi-squared test: $p<10^{-3}$), whereas no significant differences were detected for all remaining tasks (Chi-squared test: $p>0.05$).

Interestingly, among all prompt variants, data-text prompting was the only prompt variant that managed to solve the Coin Sums and Luhn tasks.
Additionally, data-text prompting significantly outperformed the other prompt variants on both of these task (Chi-squared test: $p<10^{-1}$)

Overall, these results indicate that the type of prompt provided to GPT-4o substantially influences its effectiveness in program synthesis. 
In particular, the data-text-only prompt should be preferred when the necessary information to construct it is available.
This preference is justified by the higher number of PSB2 tasks successfully solved using data-text prompting, and the fact that no other prompting variant demonstrated a statistically significant improvement over it.

\subsection{Results for success rates with 50 training cases}

\begin{table}[ht!]
\centering
  \caption{Success Rates for PSB2 tasks for PushGP and GPT-4o when provided 50 training cases.
  The underlined numbers indicate the highest success rates in a given row.
  (D) represents data-only prompts, (T) represents text-only prompts, and (D-T) represents the combination of both prompts.
  The last row shows the number of tasks solved by a particular method that were not solved by other methods.
  We discuss specific p-values of pairwise comparisons in the text.}
  \label{tab:succ_rate_50}
  \begin{tabular}{l|c|c|c|c}
    \toprule
     Task & PushGP & GPT-4o (D)  & GPT-4o (D-T) & GPT-4o (T)\\
       \midrule
        Basement & 0 & 0 & \underline{100} & \underline{100} \\
        Bouncing Balls & 0 & 0 & 0 & 0 \\
        Bowling & 0 & 12 & \underline{26} & 7 \\
        Camel Case & 0 & 98 & \underline{100} & \underline{100} \\
        Coin Sums & 13 & 0 & \underline{74} & 0 \\
        Cut Vector & 0 & 0 & 0 & 0 \\
        Dice Game & \underline{2} & 0 & 0 & 0 \\
        Find Pair & 6 & 50 & \underline{100} & \underline{100} \\
        Fizz Buzz & 66 & \underline{100} & \underline{100} & \underline{100} \\
        Fuel Cost & 63 & 2 & \underline{100} & \underline{100} \\
        GCD & 9 & 96 & \underline{100} & \underline{100} \\
        Indices of Substring & \underline{1} & 0 & 0 & 0 \\
        Leaders & 0 & 0 & 0 & 0 \\
        Luhn & 0 & 0 & \underline{2} & 0 \\
        Mastermind & 0 & 34 & \underline{77} & 0 \\
        Middle Character & 70 & 34 & 98 & \underline{100} \\
        Paired Digits & 16 & 12 & \underline{91} & 67 \\
        Shopping List & 0 & 2 & \underline{56} & 0 \\
        Snow Day & \underline{1} & 0 & 0 & 0 \\
        Solve Boolean & 1 & 0 & 19 & \underline{20} \\
        Spin Words & 0 & 80 & \underline{100} & \underline{100} \\
        Square Digits & 15 & 60 & \underline{100} & \underline{100} \\
        Substitution Cipher & 80 & 41 & 98 & \underline{100} \\
        Twitter & \underline{42} & 22 & 1 & 0 \\
        Vector Distance & 0 & 0 & 0 & 0 \\
        \midrule
        Num. Tasks Solved & 14 & 14 & 18 & 13\\
        \midrule
        Num. of Unique Tasks Solved & 3 & 0 & 1 & 0 \\
  \bottomrule
\end{tabular} 
\end{table}

\subsubsection{GPT-4o with data-text solves the most tasks with 50 training cases}


As shown in Table \ref{tab:succ_rate_50}, program synthesizers that incorporated task descriptions as part of the user intent did not exhibit a reduction in the number of tasks solved with the decreased number of training cases.
Specifically, GPT-4o achieved consistent performance across both data-text and text-only prompting conditions, solving 18 and 13 tasks, respectively.
Notably, the set of tasks solved remained identical across both prompting variants and across different numbers of provided training examples.
These results suggest that both data-text and text-only prompting exhibit robust problem-solving capabilities when provided with task descriptions and a limited set of input-output examples.


Program synthesizers that relied solely on input-output examples exhibited decreased problem-solving performance as the number of training instances was reduced from 200 to 50. 
Specifically, the number of tasks solved by PushGP dropped from 17 to 14, while GPT-4o under data-only prompting exhibited a slight decline from 15 to 14. 
Notably, the 14 tasks solved by GPT-4o with 50 training examples were entirely contained within the set of tasks it solved with 200 examples, with the sole exception of the Solve Boolean task that was only solved under the 200-example condition.
In contrast, of the 14 tasks solved by PushGP with 50 examples, 13 overlapped with those solved with 200 examples, with Square Digits being the only task uniquely solved in the 50-example condition. 
While the change in training cases caused a drop in the number of tasks solved by both program synthesizers that relied on input-output examples, PushGP managed to solve a task that the larger number of training cases condition could not, while GPT-4o saw no novel tasks being solved and was a complete overlap with the larger training cases.


The number of available input-output examples can influence the problem-solving performance of both PushGP and GPT-4o prompt-based approaches. 
In the case of PushGP, a reduced number of examples constrains the evaluation of candidate programs during evolutionary search, potentially degrading the quality of the fitness signal and limiting the effectiveness of selection. 
While a larger number of test cases generally improves solution quality, it also increases the computational cost associated with evaluating each candidate program.
For GPT-4o, the effect is more nuanced: an excessive number of examples may exceed the model’s context window or dilute the relevance of key information, while too few examples may fail to provide sufficient signal for effective learning and generalization.
These results, however, highlight the robustness of GPT-4o under data-text prompting, particularly when provided with a task description and a limited number of input-output examples. Notably, GPT-4o with data-text prompting also solved more PSB2 benchmark tasks than any other program synthesizer evaluated with both training case sizes.

\subsection{GPT-4o produces highly similar programs}

\begin{figure}[!th]
\centering
\includegraphics[width=\textwidth]{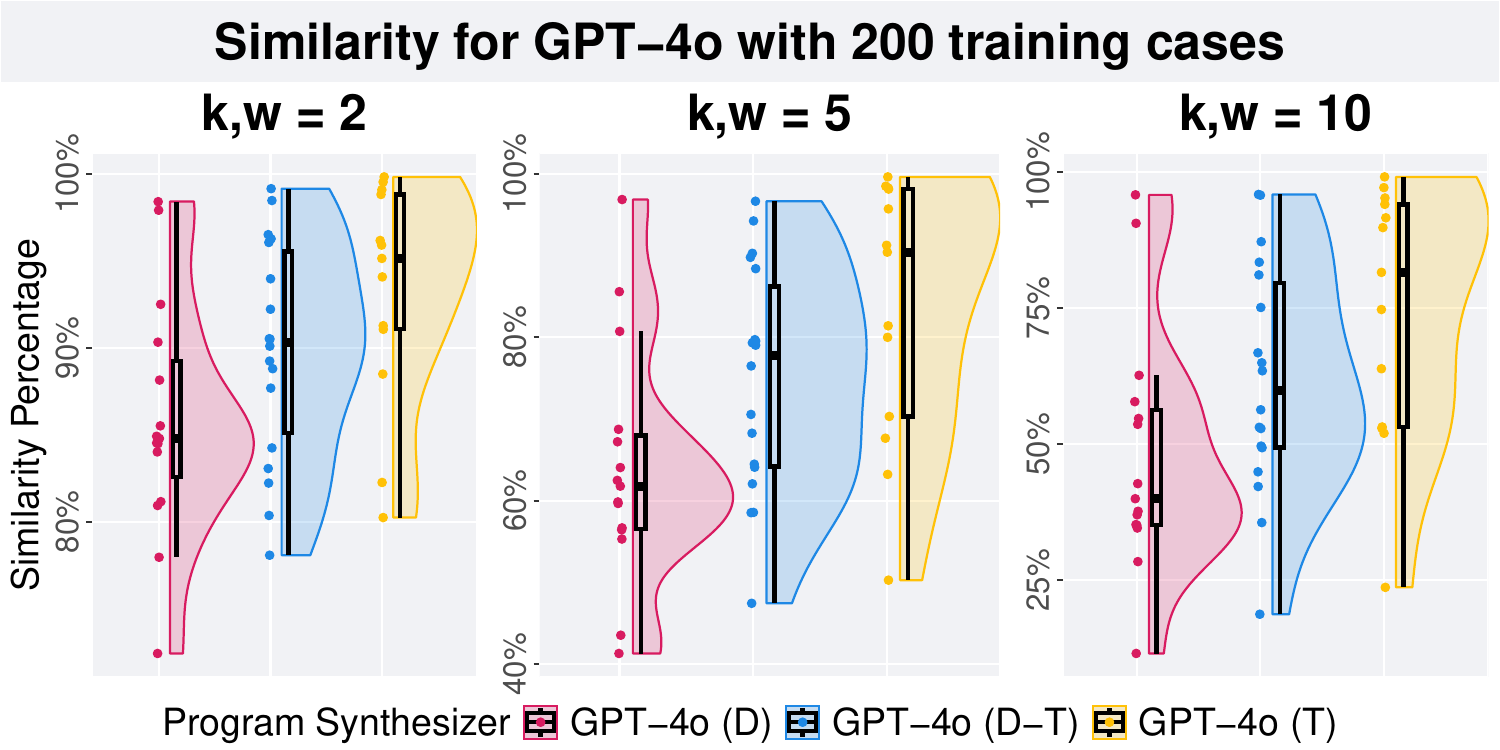}
\caption{Similarity percentage across prompt variants for solved PSB2 tasks.}
\label{fig:similarity}
\end{figure}

Figure \ref{fig:similarity} presents the average similarity percentages among successful programs synthesized by GPT-4o across the three prompt variants, for each PSB2 task with at least two successful replicates. 
To compute these similarity scores, we apply three configurations of the Winnowing algorithm \citep{schleimer2003winnowing}, where the window size (w) is set equal to the k-gram size (k), to enable a more nuanced evaluation of program similarity. 
As expected, the configuration with the smallest k-gram size (k,w = 2) yields the highest similarity percentages, since shorter k-grams are more likely to overlap. 
Conversely, with larger k-gram sizes (k,w = 5 and k,w = 10), similarity percentages decrease, reflecting the reduced likelihood of overlap among longer subsequences.

We detected significant differences among all prompt variants for each Winnowing algorithm configuration when 200 training cases are used (Kruskal-Wallis test for k,w=2: $p=0.03558$; Kruskal-Wallis test for k,w=5: $p=0.01842$; Kruskal-Wallis test for k,w=10: $p=0.0104$).
The similarity for GPT-4o with data-only prompting was significantly different than the text-only prompting across all Winnowing algorithm configurations (Wilcoxon Rank Sum test for k,w=2: $p=0.043$; Wilcoxon Rank Sum test for k,w=5: $p=0.032$; Wilcoxon Rank Sum test for k,w=10: $p=0.02$).
However, no significant differences in program similarity were observed in the remaining pairwise comparisons (i.e., comparing data-only prompting to either data-only or text-only prompting) across all configurations of the Winnowing algorithm (Wilcoxon Rank Sum test: $p > 0.05$).
No significant differences were detected for the 50 training case configuration (Kruskal-Wallis test: $p > 0.05$).

We suspect that data-only prompting leads to significantly different program similarity because GPT-4o may rely more on retrieving programs from its training corpus associated with particular test cases that are likely distributed across multiple variants of the same underlying program.
In contrast, text-only prompting appears to yield higher similarity, potentially because the task descriptions constrain GPT-4o to a narrower subset of programs that share overlapping code fragments.
A more noteworthy observation is that the program similarity under data-text prompting overlaps with both data-only and text-only variants, where it is neither significantly higher than data-only prompting nor significantly lower than text-only prompting.
Future work should investigate the sources from which GPT-4o derives its successful program outputs to gain a deeper understanding of the observed similarity patterns.

%% file: Text/conclusion.tex



\section{Conclusion}

In this work, we compare the program synthesis capabilities of PushGP and GPT-4o by evaluating their performance on tasks from the PSB2 benchmark suite \citep{helmuth2021psb2}.
Specifically, we evaluate GPT-4o with three prompt variants that capture different kinds of information that may be provided when synthesizing programs: data-only, text-only, and data-text prompting.
In total, four program synthesizers are compared.
Additionally, we vary the number of training cases available to those program synthesizers that incorporate input-output examples within their user intent.

We found that no single program synthesizer managed to consistently outperform all others with respect to solving tasks from the PSB2 suite.
However, we did find that GPT-4o with data-text prompting managed to solve the most tasks (18), regardless of training set size.
PushGP solved the largest number of unique tasks not solved by any other synthesizer: five tasks with 200 training examples and three tasks with 50 training examples.
Additionally, the high program similarities produced by GPT-4o with 200 training examples suggest a promising direction for future research, particularly in understanding the underlying factors contributing to these highly similar programs.
Ultimately, we conclude that the combination of PushGP and GPT-4o with data-text prompting leads to the greatest likelihood of generating a successful computer program, as the results show that the combination leads to the largest number of solved tasks (23 of the 25 PSB2 tasks).

